\documentclass[12pt]{article}
\usepackage{scicite}
\usepackage{graphicx}
\usepackage{times}
\usepackage{amsfonts}
\usepackage{amsmath}
\usepackage{float}
\usepackage{color, soul}
\usepackage{diagbox}
\usepackage{caption}
\usepackage{hyperref}
\usepackage{titling}
\usepackage{pdfpages}
\usepackage{amssymb} 
\usepackage{booktabs}

\usepackage{authblk}
\usepackage{longtable}

\usepackage[superscript, biblabel, nomove]{cite}

\title{Beyond Human-Like Processing: Large Language Models Perform Equivalently on Forward and Backward Scientific Text}

\author[1, *]{\mbox{Xiaoliang Luo}}
\author[2]{\mbox{Michael Ramscar}}
\author[1]{\mbox{Bradley C. Love}}
\affil[ ]{%
$^{1}$Department of Experimental Psychology, University College London, London, United Kingdom. $^{2}$Department of Psychology, University of Tübingen, Germany. *Corresponding author: xiao.luo.17@ucl.ac.uk}

\begin{document} 

\date{}
\maketitle

\begin{abstract}
The impressive performance of large language models (LLMs) has led to their consideration as models of human language processing. 
Instead, we suggest that the success of LLMs arises from the flexibility of the transformer learning architecture. To evaluate this conjecture, we trained LLMs on scientific texts that were either in a forward or backward format. Despite backward text being inconsistent with the structure of human languages, we found that LLMs performed equally well in either format on a neuroscience benchmark, eclipsing human expert performance for both forward and backward orders.
Our results are consistent with the success of transformers across diverse domains, such as weather prediction and protein design. This widespread success is attributable to LLM's ability to extract predictive patterns from any sufficiently structured input. Given their generality, we suggest caution in interpreting LLM's success in linguistic tasks as evidence for human-like mechanisms.
\end{abstract}

\section{Introduction}
Transformer-based \cite{vaswani_attention_2023} large language models (LLMs) have sparked considerable interest as models of human language processing due to their ability to generate coherent text. These models could potentially inform hypotheses about language acquisition and competence. Indeed, transformer-based language models have shown superior performance compared to recurrent language models on tasks requiring syntactic knowledge (see \cite{milliere_language_2024} for a review) such as grammaticality judgement \cite{ambridge_large_2024}, subject-verb agreement \cite{linzen_assessing_2016,goldberg_assessing_2019}, and nested recursive structure processing \cite{lakretz_mechanisms_2021,lampinen_can_2023}. Representational analyses have decoded linguistic properties from transformer models' internal representations (e.g., \cite{rogers_primer_2020,lin_open_2019}). Moreover, transformer-based language model trained on next-word prediction task can accurately capture neural and behavioral responses related to language processing \cite{schrimpf_neural_2021}.

However, it remains an open question whether LLMs are suitable models of  human language processing. In contrast to LLMs, humans learn word meanings in the context of the world \cite{arnon2012granularity}, which is not possible in language models limited to processing text. The impact of morphological generalizations on LLM performance remains unclear, despite their role in human linguistic productivity. Human multi-modal learning differs significantly from LLMs, involving unsupervised alignment of similarity structures across modalities \cite{aho_signatures_2023}, contrasting with LLMs' reliance on large-scale, matched text-image pairs \cite{alayrac_flamingo_2022,clerkin_real-world_2022}. Cognitive constraints also differentiate human and LLM learning, particularly regarding temporal limitations.

Comparing human and LLM performance is itself challenging. For example, models are frequently tested in zero-shot settings while humans receive task-specific training and instructions \cite{lampinen_can_2023}. Moreover, it can be difficult to compare model classes. While transformer-based architectures consistently outperform older RNN-based models on linguistic tasks \cite{linzen_assessing_2016,goldberg_assessing_2019}, it is unclear whether this success is attributable to transformers being more adept at language processing or better suited to modern hardware, optimization techniques and model and data sizes.

One possibility is that the mixed evidence for language models as plausible models of human language processing is that these models are general learning machines and human language is a subset of the patterns that an LLM can master. On this view, transformers' flexibility means they can converge or not converge with human capabilities depending on their training. In support of this view, transformer-based architectures exceed at many non-linguistic tasks, such as weather prediction \cite{bodnar_aurora_2024,das_decoder-only_2024}, planning \cite{song_llm-planner_2023}, molecular generation \cite{bagal_molgpt_2022}, protein design \cite{ferruz_protgpt2_2022}, code generation \cite{chen_evaluating_2021} and more.

One reason LLMs may converge with aspects of human performance is that prediction, which is LLMs' forte, plays a central role in human language learning and processing \cite{ramscar2021children, bannard2008stored} with predictive abilities emerging early in human development \cite{bannard2008stored}. Research shows that probabilistic prediction is a core part of language comprehension, evident across linguistic levels, from phonology to syntax, optimizing communication efficiency \cite{levy_expectation-based_2008}. Others show broader contextual information and top-down probabilistic guessing improves syntactic structure parsing \cite{charniak_maximum-entropy-inspired_2000}, implying language comprehension may rely on predictive mechanisms. This predictability is supported by language's inherent structure \cite{ramscar2013suffixing, dye2017functional, dye2018alternative}, a perspective echoed in both Gold's theory of language identification \cite{gold_language_1967} and aspects of Chomsky's universal grammar theory, which suggest universal structures may arise from communication's fundamental need for predictability. While language models may capture these predictive characteristics from their vast training data, this does not necessarily equate their processing with human cognition. Recent research on linguistic equivalence has yielded mixed results: some studies suggest models can learn ``impossible languages'' that humans would struggle with \cite{moro_large_2023,mitchell_priorless_2020}, while others demonstrate models learn natural languages more efficiently than unnatural ones \cite{kallini_mission_2024,papadopoulos_arrows_2024}.

In this contribution, we extend the ``impossible language'' paradigm by evaluating LLMs on neuroscience experiment prediction tasks that challenge even human experts. Specifically, we evaluated whether transformer-based language models trained on domain-specific literature (i.e. neuroscience) on next-word prediction objective using both forward and backward text (reversed at character-level) can match human experts on a benchmark testing neuroscience knowledge. Our impossible language consisted of character-level reversed neuroscience publications spanning twenty years—a learning format that departs significantly from human learning approaches and previous studies using word- or token-level reversal.

\begin{figure}
    \centering
    \includegraphics[scale=0.7]{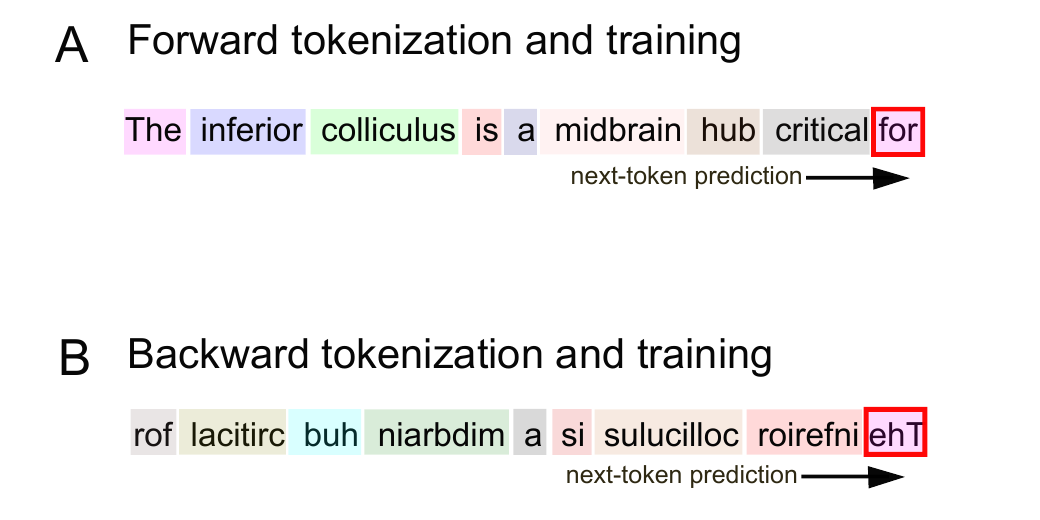}
    \caption{\textbf{Forward and backward tokenization and training.} Both forward and backward trained models were optimized to predict the next token in the training data sequence. (A) The forward tokenizer and models were trained on 20 years of neuroscience literature. (B) In contrast, the backward tokenizer and models were trained on the same data with text reversed at the character level.}
    \label{fig:fwd_vs_bwd}
\end{figure}

We adopted the GPT-2 architecture \cite{radford_language_2019} which was the foundation of many more recent LLMs such as GPT-4 \cite{openai_gpt-4_2024}, Claude \cite{anthropic_claude_2024} and Gemini \cite{team_gemini_2024}. GPT-2 is based on the transformer architecture initially trained on internet-sourced text data on a next-word prediction objective. Here, we retrained the model from scratch on twenty years of neuroscience publications. GPT-2 comes in with different sizes, we adopted three sizes of GPT-2 from small, medium to large (124M, 335M and 774M parameters). Our backward models were trained on text reversed at the character level, similar to \cite{papadopoulos_arrows_2024}. We used a dedicated tokenizer trained on the same domain-specific data (Fig. \ref{fig:fwd_vs_bwd}). This approach builds on previous work showing that specialized tokenizers can improve model performance in domain-specific tasks \cite{luo_matching_2024}. 

\begin{figure}
\centering
\includegraphics[width=\textwidth]{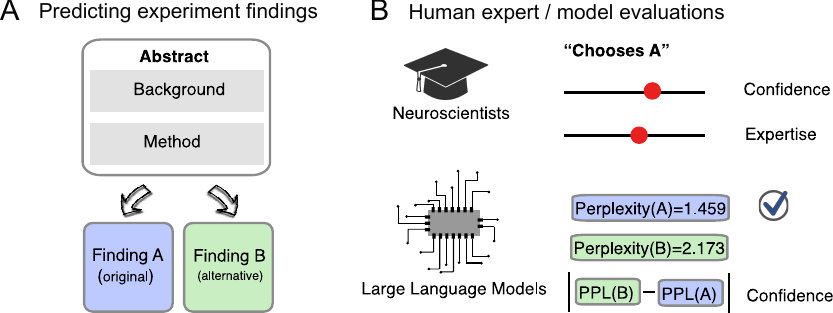}
\caption{\textbf{BrainBench is a benchmark for neuroscience.} (A) BrainBench evaluates test-takers' ability to predict neuroscience results. Test-takers chose between the original abstract and one altered to significantly change the result while maintaining coherency. (B) Human experts and Language Models (LLMs) were tasked with selecting the correct (i.e., original) version from the two options. Human experts made choices, and provided confidence and expertise ratings in an online study. LLMs were scored as choosing the abstract with the lower perplexity (i.e., the text passage that was less surprising to the model) and their confidence was proportional to the difference in perplexity between the two options. Figure adapted from \cite{luo_large_2024}.}
\label{fig:brainbench}
\end{figure}

We evaluated our models on \textit{BrainBench}, a benchmark testing predictive abilities of human experts and language models in neuroscience \cite{luo_large_2024}. BrainBench assesses how well test-takers can predict neuroscience results by presenting two versions of an abstract from a recent journal article. The task is to choose between the original and an altered version, where the latter significantly changes the study's outcome while maintaining overall coherence (Fig. \ref{fig:brainbench}A). Human experts made binary decisions and provided confidence and expertise ratings in an online study (Fig. \ref{fig:brainbench}B). Language models were scored based on choosing the version with lower perplexity (i.e., the less surprising text passage). Their confidence was proportional to the perplexity difference between the two options (Fig. \ref{fig:brainbench}B). BrainBench's test cases were sourced from recent {\em Journal of Neuroscience} abstracts across five neuroscience domains: Behavioral/Cognitive, Systems/Circuits, Neurobiology of Disease, Cellular/Molecular, and Developmental/Plasticity/Repair (Fig. \ref{fig:brainbench}C).

To foreshadow our results, remarkably, backward-trained models excelled at BrainBench problems despite training on reversed text, which violates natural language order and human learning patterns—a process many might anticipate would hinder performance. Despite higher perplexity on reversed training and testing items, backward-trained models, performed (non-significantly) better than forward-trained models on our benchmark. These models matched (GPT-2 124M) or surpassed (GPT-2 774M) human expert performance.

We conclude that language models are general pattern learning machines, capable of processing input regardless of whether it reflects human language structures. This explains their success in numerous machine learning applications as mentioned above. However, their flexibility and generality in handling different types of inputs should not be mistaken as evidence of human-like language learning processes.

\section{Results}
\subsection{Forward and Backward-Trained Language Models Are Strong BrainBench Predictors}
We evaluated GPT-2 models of varying sizes, trained both forward and backward from scratch on neuroscience literature, on the BrainBench task (selecting the correct version of an abstract) and compared them against human-expert performance from \cite{luo_large_2024}, involving 171 neuroscientists (see Methods).

\begin{figure}[H]
    \centering
    \includegraphics[scale=0.5]{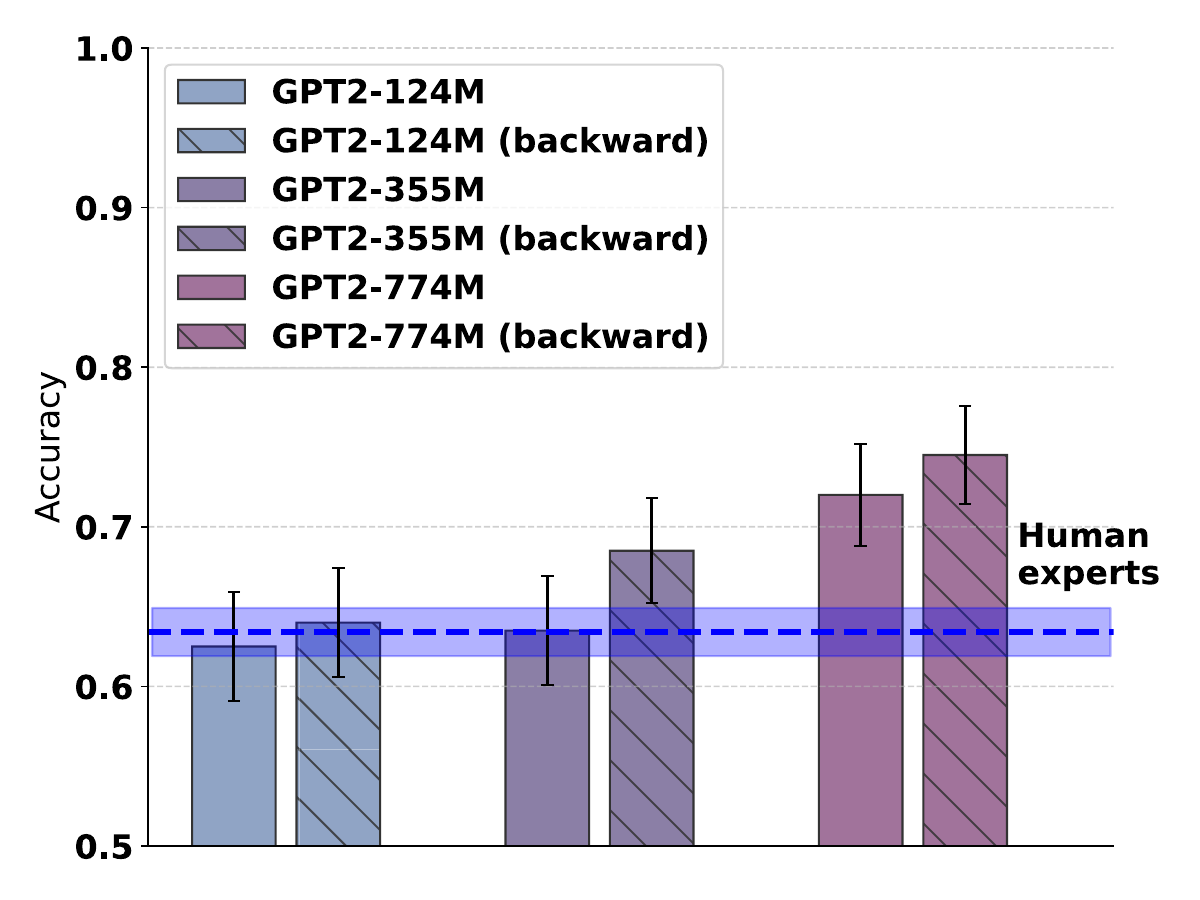}
    \caption{\textbf{BrainBench performance of GPT-2 models trained forward and backward.} GPT-2 models, trained from scratch on two decades of neuroscience literature, rival or exceed human expert performance, demarcated by the blue dashed line. Models trained on the same data reversed at the character level performed non-significantly better than their forward-trained counterparts.} 
    \label{fig:brainbench_overall_acc}
\end{figure}

Notably, backward-trained GPT-2 models were non-significantly more accurate on BrainBench than their forward-trained counterparts ($F(1, 2) = 2.77, p = .238$; Fig. \ref{fig:brainbench_overall_acc}). As model size increased, both forward and backward-trained models improved in BrainBench performance ($F(1, 2) = 20.51, p = .046$) with larger models surpassing human-level accuracy. The interaction between direction and model size was not significant ($F(1, 2) = 0.05, p = .842$). 

While the difference between forward and backward-trained models of the same size was not statistically significant, we hypothesize that their strong performance of backward-trained models might be attributed to their tokenizer generating more neuroscience-related tokens compared to the forward-trained tokenizer (Fig. \ref{fig:neuro_tokens}).

\subsection{Backward-Trained Models Exhibit Consistently Higher Perplexities Compared to Forward-Trained Models}
\begin{figure}[H]
\centering
\includegraphics[scale=0.5]{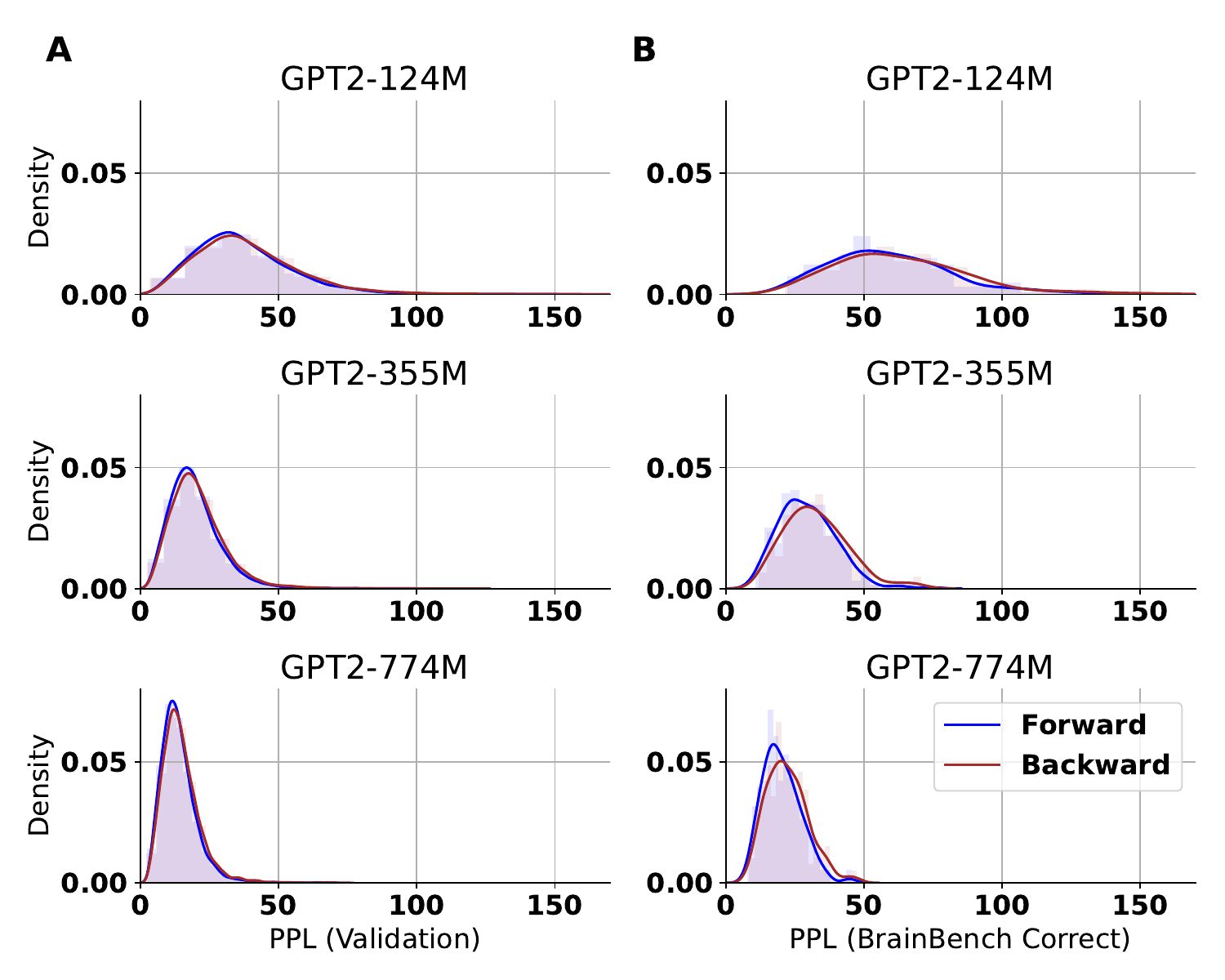}
\caption{\textbf{Backward-trained models exhibit higher perplexities on both validation and BrainBench items.} (A) Perplexity of the validation set items; (B) Perplexity of the correct options in BrainBench items.}
\label{fig:ppl_diff_val_and_test}
\end{figure}

Despite similar performance on BrainBench, we investigated potential differences in how forward and backward-trained GPT-2 models learn and process textual data. We evaluated the perplexity of both the validation data from the twenty-year neuroscience corpus and the BrainBench items.
Backward-trained GPT-2 models consistently demonstrated significantly higher perplexities on both the validation data (Fig. \ref{fig:ppl_diff_val_and_test}A; Table \ref{tab:ppl_diff_val_and_test}) and the BrainBench test items (Fig. \ref{fig:ppl_diff_val_and_test}B; Table \ref{tab:ppl_diff_val_and_test}). This pattern held across all model sizes and datasets tested.

\begin{table}[H]
\centering
\begin{tabular}{lcc}
\toprule
\textbf{Model Size} & \textbf{PPL (Validation)} & \textbf{PPL (BrainBench Correct)}\\
\midrule
124M & $t(18767) = -6.101, p = 0.000$ & $t(199) = -9.368, p = 0.000$\\
355M & $t(18767) = -7.609, p = 0.000$ & $t(199) = -14.509, p = 0.000$\\
774M & $t(18767) = -8.194, p = 0.000$ & $t(199) = -14.740, p = 0.000$\\
\bottomrule
\end{tabular}
\caption{\textbf{Backward-trained models exhibit higher perplexities on validation and BrainBench items.} Across model sizes, backward-trained models achieve higher perplexities on both validation set items and BrainBench test items than forward-trained models.}
\label{tab:ppl_diff_val_and_test}
\end{table}

\subsection{Backward-Trained Models Are Less Aligned with Human Judgements}
We investigated whether forward-trained and backward-trained models, as well as human experts, found the same BrainBench items challenging. For humans, we calculated the mean accuracy for each of the 200 test cases. For GPT-2 models, we computed the signed differences in perplexity between incorrect and correct abstracts for each test case (see Methods).
Our analysis revealed that model judgments, regardless of training direction, generally correlate more strongly with each other ($M=0.69, SD=0.07$) than with human judgments ($M=0.09, SD=0.04$). While the overall correlation between model and human expert judgments is low, backward-trained models showed significantly lower correlation ($M=0.05, SD=0.02$) to human judgments compared to forward-trained models ($M=0.13, SD=0.01$; $t(2) = 20.848, p = 0.002$). 


\begin{figure}[H]
    \centering
    \includegraphics[scale=0.3]{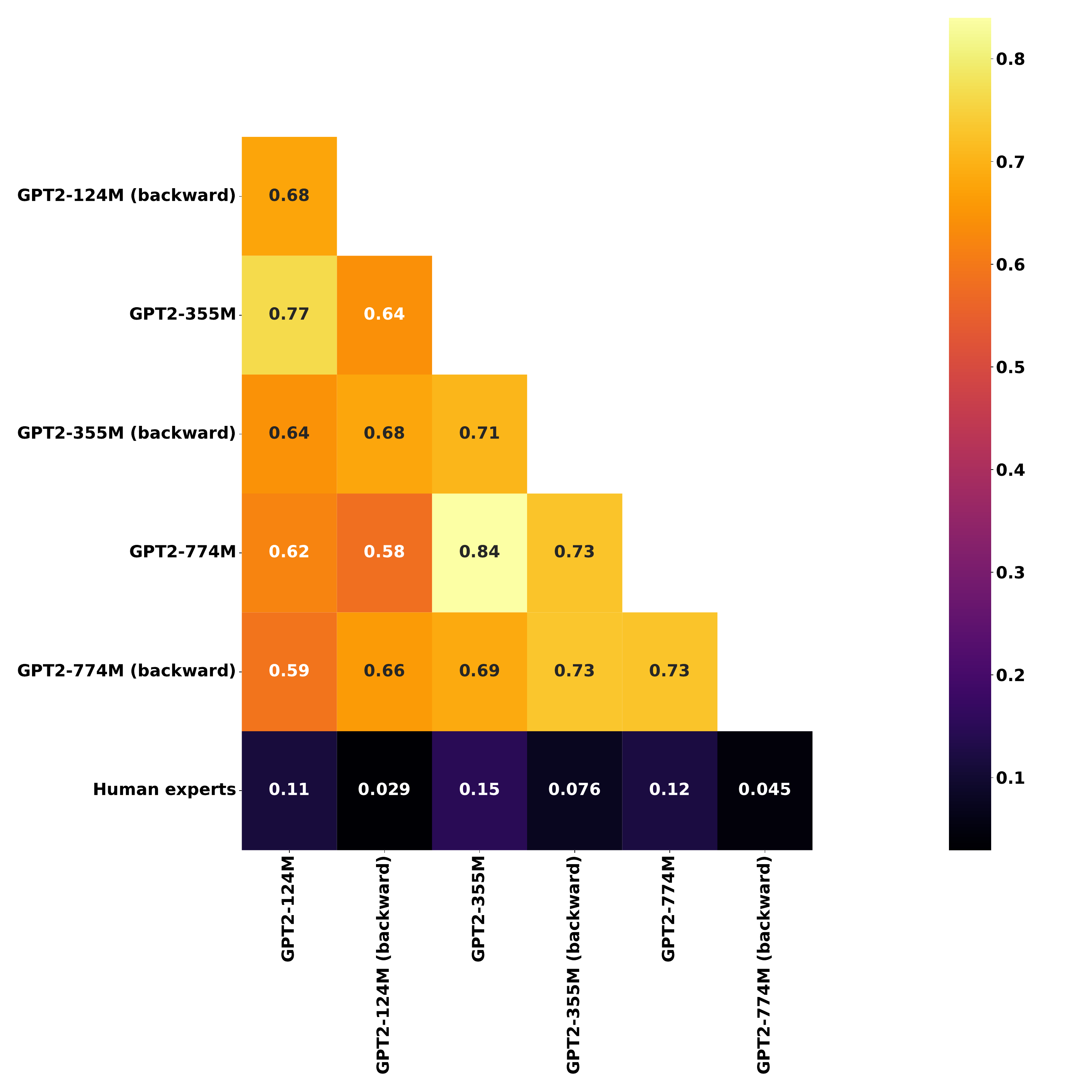}
    \caption{\textbf{Comparison of model and human judgments on BrainBench difficulty.} Model judgments (both forward and backward-trained) correlate more strongly with each other than with human expert judgments. Backward-trained models show significantly lower correlation to human judgments compared to forward-trained models.}
    \label{fig:error_correlation}
\end{figure}

\section{Discussion}
We demonstrated in this contribution that transformer-based language models can predict neuroscience experimental outcomes regardless of whether they are trained on forward or backward text. Remarkably, models trained on character-level reversed text performed numerically (but not significantly) better than their forward-trained counterparts on the BrainBench task. Both forward and backward-trained models matched or exceeded human expert performance as model size increased, with the largest models (774M parameters) showing superior performance to human experts. Backward-trained models exhibited higher perplexities than forward-trained models on the validation split of the twenty-year neuroscience corpus and the BrainBench items. In addition, forward-trained models correlated better with expert judgments of BrainBench item difficulty than backward-trained models, though both showed weaker human correlation compared to inter-model agreement.

These findings suggest that transformer-based language models are best understood as general pattern learning machines rather than specific models of human language processing. The success of backward-trained models, despite operating on input that violates fundamental properties of human language, indicates that these models can extract predictive structure from their training data regardless of its format or alignment with human cognitive constraints. While higher perplexity scores suggest that backward-trained models find the reversed text more surprising or uncertain, this additional uncertainty does not impair their ability to discriminate between correct and incorrect experimental outcomes in BrainBench. The higher perplexity aligns with how human languages naturally evolved—they are optimized for forward processing in real-time conversation, where listeners must interpret speech sequentially. Since human cognitive limitations shaped language to be processed primarily in a forward direction, it is expected that models would show higher uncertainty when processing reversed text while still maintaining their core ability to extract meaning.

Prior studies examining reversed text in language models \cite{mitchell_priorless_2020,kallini_mission_2024} have primarily reversed text at word or token boundaries while retaining the original tokenizer. Our approach differs substantially where we trained a new tokenizer from scratch on character-reversed neuroscience text (also see \cite{papadopoulos_arrows_2024}). Moreover. our study uniquely evaluates these models on specialized tasks that challenge human experts. Our findings serve as a crucial reminder against drawing simplistic parallels between transformer architectures and neural processing in the brain. The ability of LLMs to achieve superhuman performance on specialized tasks while processing extreme input formats—impossible for human learning—highlights fundamental differences in their underlying capabilities and constraints. Unlike LLMs, human language processing is shaped by specific cognitive constraints, multimodal experiences, and interactive social contexts.

This is not to dismiss LLMs as mere ``stochastic parrots'' \cite{bender_dangers_2021}—their ability to extract and leverage complex predictive patterns from diverse input formats demonstrates sophisticated learning capabilities. However, their very generality and flexibility, which enables impressive performance across domains, makes them problematic models of human cognition. When LLMs diverge from human-like behaviour, it may reflect their fewer inherent constraints rather than their limitations. They succeed wherever sufficient predictive structure exists, regardless of its alignment with human cognitive architectures or natural language properties.

\newpage
\section{Methods}
\subsection{BrainBench}
BrainBench \cite{luo_large_2024} is a benchmark consists of 200 test cases from abstracts in the \textit{Journal of Neuroscience published in 2023}. These abstracts are categorized into five sections: Behavioral/Cognitive, Systems/Circuits, Neurobiology of Disease, Development/Plasticity/Repair, and Cellular/Molecular.

Each test case contains a published abstract and an altered version crafted by neuroscientists (see details in \cite{luo_large_2024}). These modifications, though minimal, significantly change the results—for instance, by changing the roles of brain regions or reversing a result's direction (e.g., from "decreases" to "increases"). The altered abstracts remain logically coherent despite the changes.

The BrainBench task is to identify the correct study outcome by choosing between the original abstract and its altered counterpart.

\subsection{Model evaluation}
Two versions of the abstracts from each test case were presented to models separately. We prefixed each abstract with the prompt ``\textit{You are a neuroscientist with deep knowledge in neuroscience. Here is an abstract from a neuroscience publication:}''. We then measured the perplexity of both passages and used perplexity as the indicator of whether models favor one abstract or the other.

Perplexity measures the degree of uncertainty of a model when generating a particular sequence of text and is defined as the exponentiated average negative log-likelihood of a tokenized sequence. If we have a tokenized abstract \( X = (x_0, x_1, \ldots, x_t) \), then the perplexity of \( X \), given a model parameterized by $\theta$ is,

\begin{equation}
    PPL(X) = \exp \left\{ -\frac{1}{t} \sum_{i}^{t} \log p_\theta (x_i | x_{<i}) \right\}
\label{eq:ppl}
\end{equation}
where \( \log p_\theta (x_i | x_{<i}) \) is the log-likelihood of the \( i \)th token conditioned on the preceding tokens \( x_{<i} \) according to the model. Given both the original and the altered abstracts, we used the abstract with lower perplexity as the model's decision and evaluated the overall accuracy across the entire BrainBench dataset accordingly.

\subsection{Human evaluation}
Previous work \cite{luo_large_2024} collected human judgements from 171 neuroscience experts on BrainBench. These data are publicly available\footnote{https://github.com/braingpt-lovelab/BrainBench} and provide a useful comparison to LLM performance.

\subsection{Error correlation}
To assess performance correlation across LLMs and between humans and LLMs, we employed a difficulty-based approach. For LLMs, difficulty was measured by the perplexity difference between incorrect and correct abstracts for each test case, with larger positive differences indicating easier cases. Human difficulty was determined by mean accuracy per item. We then calculated Spearman correlation coefficients between these difficulty measures to evaluate agreement among LLMs and between LLMs and human experts, providing insight into how different models and humans rank the relative challenges of test cases.

\subsection{Training data}
\label{training_data}
The data we used to train the GPT-2 variants from scratch were collected by \cite{luo_large_2024}. The training data spans Neuroscience publication (abstracts and full articles) dates 2002-2022, totaling
1.3 billion tokens. We randomly allocated 90\% of the data for training, reserving the remaining 10\% for validation. 

\subsection{Tokenization}
\label{tokenization}
The neuro-tokenizer employs GPT-2's tokenization strategy \cite{radford_language_2019}, adapted from Byte Pair Encoding (BPE) \cite{gage_new_1994} for word segmentation \cite{sennrich_neural_2016}. It's trained anew on neuroscience data used for model training, maintaining a vocabulary size of 50,257 tokens. 

\subsection{Training details}
We trained variants of GPT-2 models using Huggingface implementations. We used a batch size of 16 for GPT-2 124M (8 for GPT-2 355M and 4 for GPT-2 774M) and a chunk size of 1024. Training involved the use of the AdamW optimizer \cite{loshchilov_decoupled_2019} with a learning rate of 2e-5 and a cosine learning rate scheduler (i.e., learning rate decays following a cosine schedule over training epochs). We applied gradient accumulation steps set at 8. Five training epochs were performed, along with a warm-up step of 0.03 and a weight decay rate of 0.001. bf16 mixed precision training and data parallelism were employed. We used 4 Nvidia A100 (80GB) GPUs hosted on Microsoft Azure.

\subsection{Backward model training and evaluation}
The backward GPT-2 models we trained from scratch follow the same procedures outlined above for tokenization, training, and evaluation. The only difference is that all texts were reversed at the character-level. This means that each training sequence, tokenizer input, and BrainBench test case had its characters reversed, where the original sequence's start became the end and vice versa.

\subsection{Statistical testing}
To test the effects of model size and training direction on prediction accuracy, we conducted a repeated-measures Analysis of Variance (ANOVA). The dependent variable was prediction correctness for each BrainBench item. Model size and direction were included as fixed factors, with model size coded as a continuous variable and direction binary-coded as a categorical variable. Model was treated as a within-subjects factor to account for repeated measurements. The analysis was implemented using the \texttt{aov()} function in \textsf{R}, with the Error term specified to accommodate the repeated-measures design. Both main effects and the interaction between model size and direction were examined.

\newpage
\section{Supplementary Information}
\renewcommand\thefigure{S.\arabic{figure}}  
\setcounter{figure}{0}  
\renewcommand{\thetable}{S.\arabic{table}}
\setcounter{table}{0}

\subsection{Neuro-token analysis of forward- and backward-trained tokenizers}
\begin{figure}[H]
    \centering
    \includegraphics[scale=0.5]{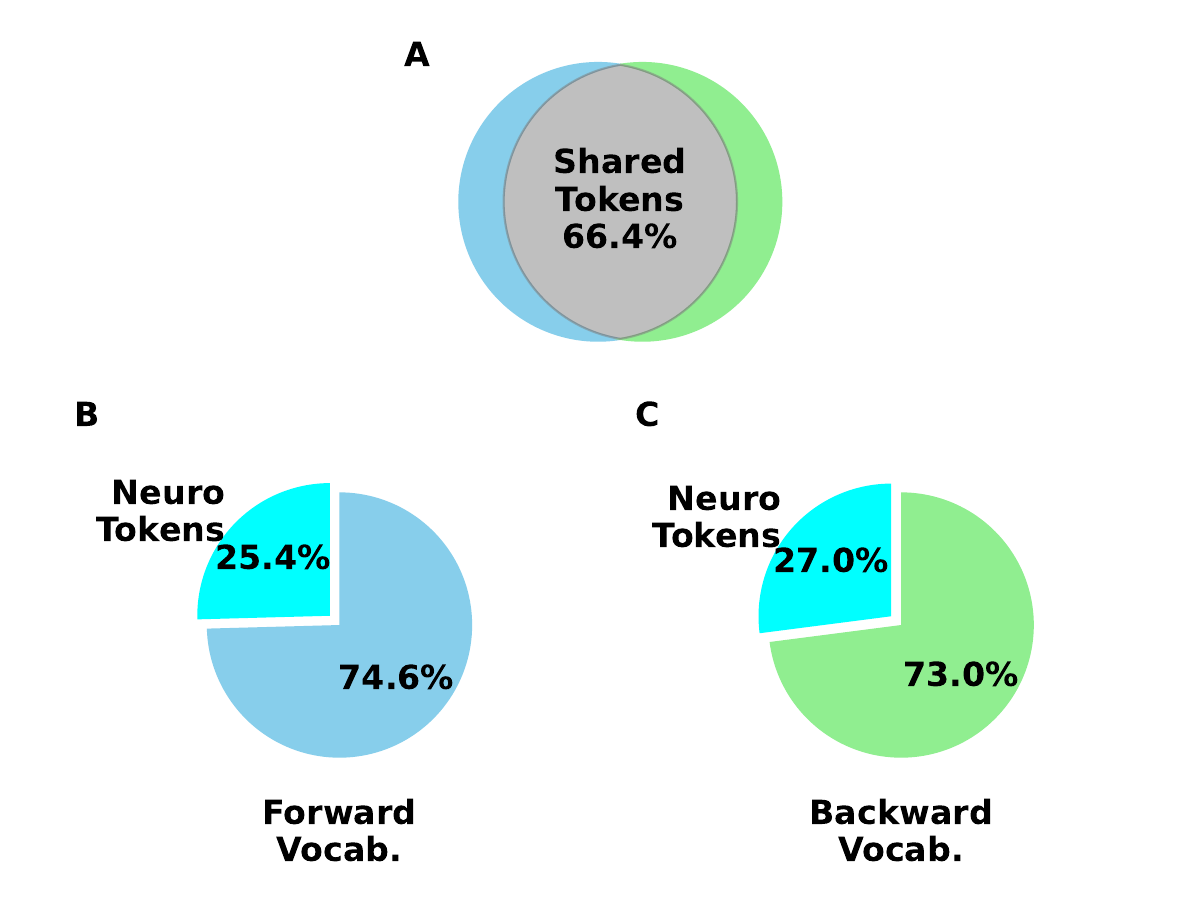}
    \caption{\textbf{Tokenizer trained on forwards vs. backwards neuroscience text results in differences in vocabularies.} (A) The forwards and backwards tokenizers share $66.4\%$ of the vocabularies once the tokens in the backwards tokenizer is reversed at character level. (B-C) Of the vocabularies from the two tokenizers, 25.4\% of the tokens from the forwards tokenizer are commonly associated with neuroscience according to GPT-4, compared to 27\% of the tokens from the tokenizer trained on backwards text.}
    \label{fig:neuro_tokens}
\end{figure}

Given both vocabularies of forward- and backward-trained tokenziers, we prompted GPT-4 (API version: 2024-02-15-preview) to categorize whether each unique token is a commonly seen term associated with Neuroscience. As shown in Fig. \ref{fig:neuro_tokens}, while both tokenizers share $66.4\%$ tokens out of a fixed vocabulary size of 50,257, the backward-trained tokenizer produced more ($27\%$) tokens associated with Neuroscience than the forward-trained tokenizer ($25.4\%$).

\bibliographystyle{naturemag}
\bibliography{references-ken,references-michael}
\end{document}